\documentclass[sigconf]{acmart}
\renewcommand\footnotetextcopyrightpermission[1]{} 

\usepackage{epsfig}
\usepackage{graphicx}
\usepackage{amsmath}
\usepackage{amssymb}
\usepackage{enumitem}

\usepackage[font=footnotesize,labelfont=bf]{caption}

\usepackage{multirow}
\usepackage{tabularx}
\usepackage{dsfont}
\usepackage{bm}
\usepackage{url}
\usepackage[tight]{subfigure}
\usepackage{color}
\usepackage{subfigure}
\usepackage{amsthm}
\usepackage[export]{adjustbox}
\usepackage[rightcaption]{sidecap}
\usepackage{comment}
\usepackage{hyperref}
\usepackage{colortbl}

\usepackage[utf8]{inputenc} 
\usepackage[T1]{fontenc}    
\usepackage{url}            
\usepackage{booktabs}       
\usepackage{amsfonts}       
\usepackage{nicefrac}       
\usepackage{microtype}      

\usepackage{balance}
\usepackage[rightcaption]{sidecap}

\def\minimize{\operatornamewithlimits{minimize}}

\definecolor{dg}{rgb}{0,0.694,0.298}
\definecolor{purple}{rgb}{0.4,0.176,0.569}

\usepackage{pifont}
\usepackage{xspace}
\makeatletter
\DeclareRobustCommand\onedot{\futurelet\@let@token\@onedot}
\def\@onedot{\ifx\@let@token.\else.\null\fi\xspace}
\def\eg{\emph{e.g}\onedot} 
\def\ie{\emph{i.e}\onedot} 
 
\def\etc{\emph{etc}\onedot} 
 
\def\etal{\emph{et al}\onedot}
\makeatother

\AtBeginDocument{%
  \providecommand\BibTeX{{%
    \normalfont B\kern-0.5em{\scshape i\kern-0.25em b}\kern-0.8em\TeX}}}

\acmConference[ ]{ }{ }

\begin{document}

\title{\emph{FakePolisher}: Making DeepFakes More Detection-Evasive by Shallow Reconstruction}



\author{Yihao Huang$^{1}$, \ Felix Juefei-Xu$^{2}$, \ Run Wang$^{3,*}$, \ Qing Guo$^{3}$, \ Lei Ma$^{4}$, \ Xiaofei Xie$^{3}$, \ Jianwen Li$^{1}$, \ Weikai Miao$^{1}$, \ Yang Liu$^{3,5}$, \ Geguang Pu$^{1,*}$}
\thanks{
Yihao Huang's email: huangyihao22@gmail.com \\
$^*$ Corresponding authors. E-mail:~{runwang1991@gmail.com, ggpu@sei.ecnu.edu.cn}}
\affiliation{\institution{$^{1}$East China Normal University, China \ \ $^{2}$Alibaba Group, USA \ \ $^{3}$Nanyang Technological University, Singapore}}
\affiliation{\institution{$^{4}$Kyushu University, Japan \ \ $^{5}$Zhejiang University, China}}
\renewcommand{\shortauthors}{Yihao Huang, et al.}

\begin{abstract}
At this moment, GAN-based image generation methods are still imperfect, whose upsampling design has limitations in leaving some certain artifact patterns in the synthesized image. Such artifact patterns can be easily exploited (by recent methods) for difference detection of real and GAN-synthesized images. However, the existing detection methods put much emphasis on the artifact patterns, which can become futile if such artifact patterns were reduced.

Towards reducing the artifacts in the synthesized images, in this paper, we devise a simple yet powerful approach termed FakePolisher that performs shallow reconstruction of fake images through a learned linear dictionary, intending to effectively and efficiently reduce the artifacts introduced during image synthesis. In particular, we first train a dictionary model to capture the patterns of real images. Based on this dictionary, we seek the representation of DeepFake images in a low dimensional subspace through linear projection or sparse coding. Then, we are able to perform shallow reconstruction of the ‘fake-free’ version of the DeepFake image, which largely reduces the artifact patterns DeepFake introduces.
The comprehensive evaluation on 3 state-of-the-art DeepFake detection methods and fake images generated by 16 popular GAN-based fake image generation techniques, demonstrates the effectiveness of our technique. Overall, through reducing artifact patterns, our technique significantly reduces the accuracy of the 3 state-of-the-art fake image detection methods, \ie, 47\% on average and up to 93\% in the worst case.

Our results confirm the limitation of current fake detection methods and calls the attention of DeepFake researchers and practitioners for more general-purpose fake detection techniques.


\begin{figure}[]
	\centering 
	\includegraphics[width=\columnwidth]{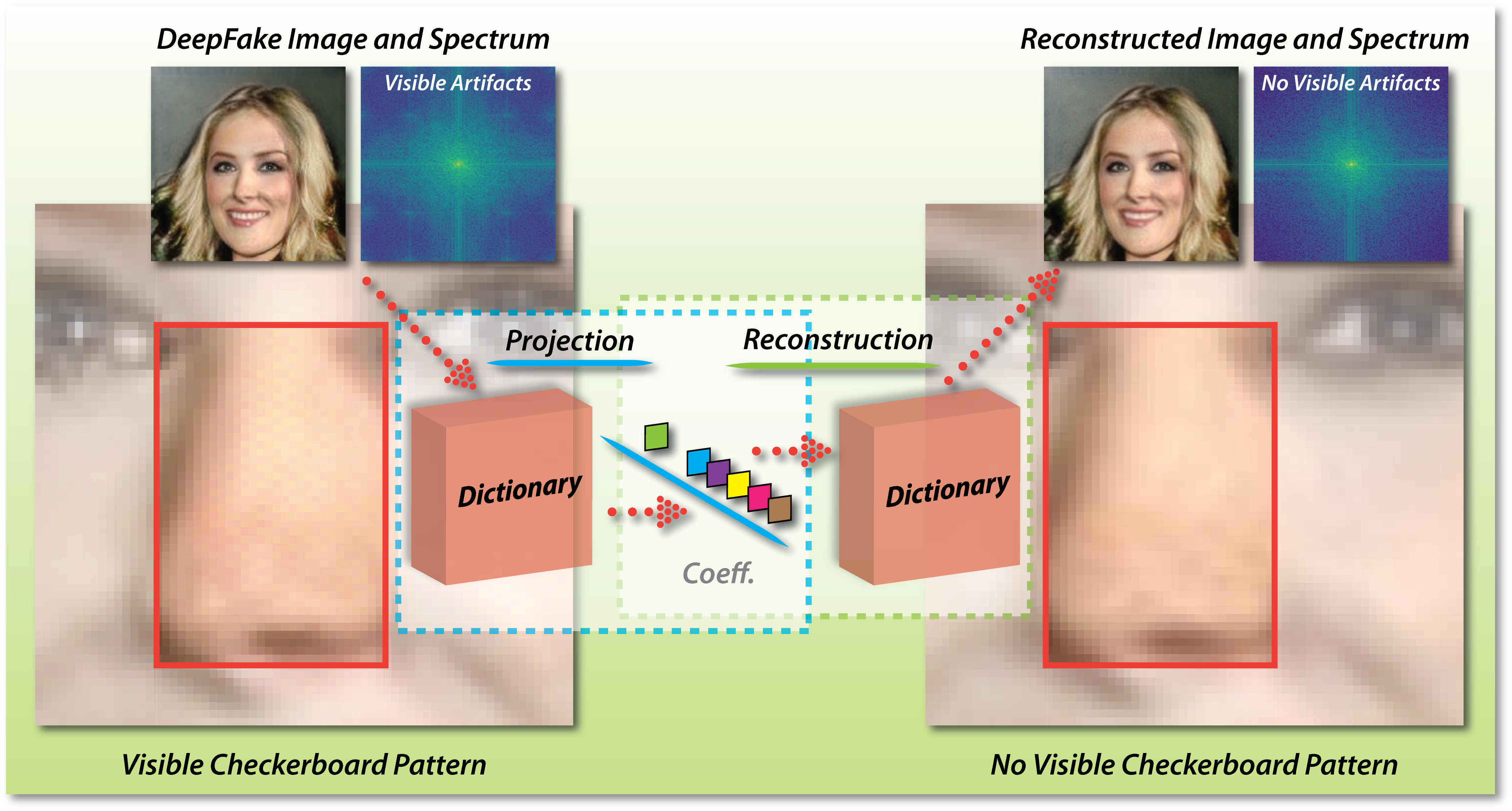}
	\caption{Before and after \emph{FakePolisher} is applied: the left image is a fake image generated from the DeepFake method \cite{choi2018stargan}. In the enlarged view, we can easily find obvious checkerboard patterns. Corresponding to these artifacts are the bright blobs at 1/4 and 3/4 of the width/height in the spectrum of the fake image. The artifacts are introduced by the upsampling methods of GAN-based image generation methods. We propose a shallow reconstruction method based on dictionary learning to remove the artifacts. The right image is the reconstructed image, which does not has obvious artifact in its enlarged view and spectrum.}
	\label{fig:teaser}
\end{figure}
\end{abstract}





\maketitle

\section{Introduction}\label{sec:intro}
The recent advances of fake information generation draw lots of attention and concern, with frequent and widespread media coverage and argument. Up to the present, \textbf{DeepFake} (\eg, fake images, audios and videos) has become a real threat to our society due to its realism and impact scopes. Even worse, lots of tools such as FaceApp \cite{FaceApp}, ZAO \cite{ZAO} are available for fake image generation, further exacerbating the situation.
In general, the backend techniques of DeepFake are mostly based on generative adversarial networks (GANs), which are used for synthesizing facial images and voices. Many of the current state-of-the-art DeepFake techniques reach a level that cannot be easily captured by human perceptions. For example, it can be really hard for humans to distinguish the real videos from faked ones only by our eyes and ears \cite{rossler2019faceforensics++,todisco2019asvspoof}. We are entering an era where we cannot simply trust our eyes and ears. According to \cite{whichfaceisreal}, humans detection peak accuracy reaches only 75\% \cite{humanperformance}, where the real images are from a well-known dataset Flicker-Faces-HQ (FFHQ) and the fake images are generated by StyleGAN \cite{karras2019style}.

Although easily fooling the human, the state-of-the-art synthesized images can still be detected in many cases by current fake detection methods. The state-of-the-art synthesized methods often introduce artifact patterns into the image during generation, opening a chance for fake detectors \cite{zhang2019detecting, frank2020leveraging}.
Due to the current technical limitation, even worse, the image manipulation footprint will be inevitably left in a synthesized image, either by partial image manipulation \cite{choi2018stargan,he2019attgan,liu2019stgan} or full image synthesis \cite{karras2017progressive,karras2019style,karras2019analyzing}.
In particular, the partial image manipulation methods often use convolutional and pooling layers to transform a real image into feature maps. After feature map modification, they have to use the upsampling method in the decoder to amplify the feature maps into a high-resolution fake image. Similarly, full image synthesis takes a random vector and amplifies it with the decoder. 
Such inevitable manipulation footprints leave traces for automated fake detection.

Thus far, most state-of-the-art fake image detection methods are proposed based on convolutional neural network (CNN), which roughly fall into three categories by their input feature types, \ie, image-based methods \cite{wang2019cnn,afchar2018mesonet,nguyen2019capsule,nguyen2019multi}, fingerprint-based methods \cite{yu2019attributing}, and spectrum-based methods \cite{zhang2019detecting,frank2020leveraging}.
\begin{itemize}[leftmargin=*]
    \item \emph{Image-based methods} adopt large and complex networks to perform fake detection, by directly working on the images as inputs.
    \item \emph{Fingerprint-based methods} leverage both fingerprints of GAN and images as inputs for fake detection, based on the assumption that GANs carry certain model fingerprints, leaving stable fingerprints in their generated images. They even possibly allow us to identify which kind of DeepFake method is used for generation.
    \item \emph{Spectrum-based methods} find out that all GAN
    architectures in the generation process leave some footprints in the frequency domain. Therefore, they propose to leverage the spectrum information for fake detection.
\end{itemize}
These three types of methods \cite{yu2019attributing,wang2019cnn,frank2020leveraging} all demonstrate their usefulness, in achieving the state-of-the-art performance for GAN-synthesized fake image detection.  


We can see that the manipulation footprints during the synthesizing process open the chance for fake image detection. Existing techniques can possibly leverage such information, from different perspectives to different extent. Therefore, a new methodology that reduces the footprint introduced during the synthesized process could increase the chance of bypassing the fake detectors.

\begin{figure}[tbp]
	\centering 
	\includegraphics[width=\columnwidth]{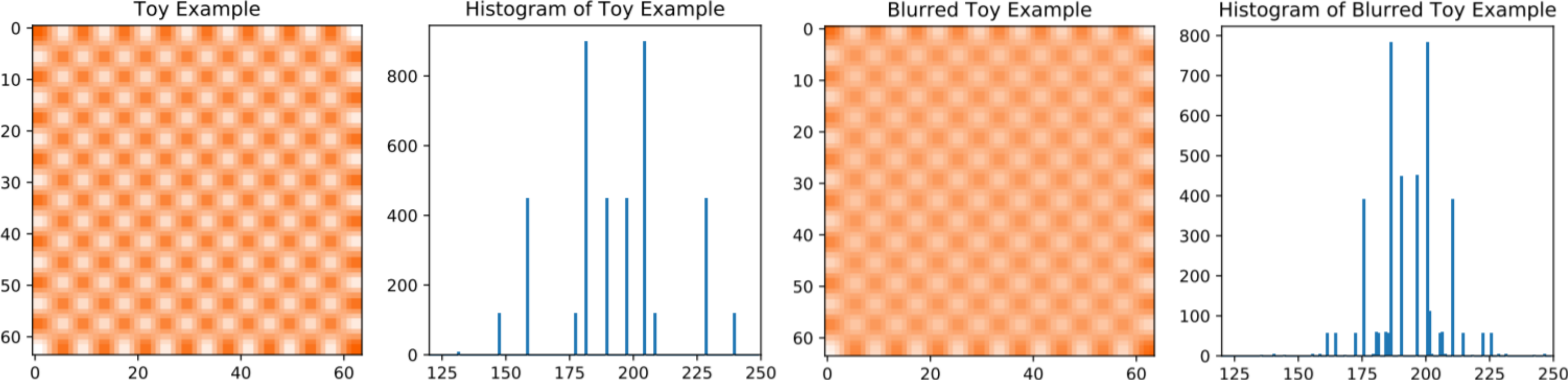}
	\caption{From left to right, the original toy example and its histogram are shown, followed by the blurred toy example and its histogram. The original toy example contains obvious checkerboard patterns. The checkerboard patterns also exist in the blurred toy example. We can find that the blur method can not effectively remove the checkerboard patterns.}
	\label{fig:toy_example}
\end{figure}

Although smoothing could be a possible way for fake footprint reduction, we find that it is generally infeasible to effectively reduce the footprint.
For example, in Figure \ref{fig:toy_example}, the first image is a toy case with checkerboard patterns, generated by the following procedures. We first produce a checkerboard image with size 8*8. The checkerboard has two colors: white and orange. Then, we resize the image to 64*64 by using interpolation, which simulates the operation of upsampling. The histogram of it calculates the distribution of the values in the gray-scale version of the toy example. The third image is the blurred toy example. We apply a 5*5 kernel of Gaussian blur to the toy example to obtain this image. Although the checkerboard patterns are weakened, they still exist in the image. In the histogram of the gray-scale blurred toy example, we can also find that the values of peaks are regular and symmetric as that in the histogram of the toy example. The features can be easily detected.

In this paper, we propose the \emph{FakePolisher}, a shallow reconstruction method with dictionary learning \cite{ksvd,pca} to reduce such fake footprints. In particular, we try to find the `closest' representation, free of fake patterns, of its fake image counterparts.
We first train a dictionary model to systematically capture the patterns of real images, based on which we seek the representation of DeepFake images in a low dimensional subspace through linear projection or sparse coding. 
Then, we perform shallow reconstruction of the ‘fake-free’ version of the DeepFake image, intending to largely reduce the manipulation footprints the DeepFake introduces.
Our in-depth evaluation on 3 state-of-the-art DeepFake detection methods and fake images generated by 16 GAN-based methods demonstrates that 
our reconstructed images successfully fool all the three types of fake image detection methods. 
Although previous work does not explicitly mention they leverage manipulation footprint for fake image detection, our method successfully reduces the accuracy of the 3 state-of-the-art detection techniques significantly with an average accuracy decrease of 47\%. This indicates that existing fake detection methods highly relies on the manipulation footprint introduced in the synthesizing phase.
Our study presents a new challenge for future fake image detection methods in the domain of multimedia forensics, which need to look for more advanced fake patterns beyond only the footprints introduced by the generation phase. 




The main contributions of this paper are summarized as follows.
\begin{itemize}[leftmargin=*]
    \item To reduce the footprint in GAN-synthesized fake images, we propose a post-processing shallow reconstruction method by using dictionary learning, which does not rely on any information of the GAN used for generation.
    In other words, it can be used as a black-box attack method to fool the fake image detectors.
    \item We conduct a comprehensive evaluation of our proposed approach in fooling three representative state-of-the-art of fake image detection methods over fake images generated by 16 GAN-based methods. 
    By reducing the manipulation footprints, our method reduces the fake detection accuracy of these methods significantly.
    Our reconstructed images also exhibit high similarity to its original fake image counterpart.
    \item So far, it is still unknown whether existing
    fake detection methods leverage the manipulation footprints and in what ways. Our results answer this question, indicating that existing methods can highly leverage the manipulation of footprint information from different perspectives. Our results call for attention that more general fake detection mechanisms should be designed.
\end{itemize}

\section{Related Work}\label{sec:related}
Since its advent, GAN \cite{goodfellow2014generative} has been successfully applied to many application domains, especially in the generation process for images, natural languages, and audios, \etc

\subsection{GAN-based Image Generation}
Over the past several years, a lot of GAN-based image generation methods have been proposed, largely following two categories: full image synthesis and partial image manipulation. 

\noindent\emph{Full image synthesis.} Progressive growing GAN (ProGAN) \cite{karras2017progressive} is able to synthesize high-resolution images via the incremental enhancement of the discriminator and the generator networks during the training process. StyleGAN \cite{karras2019style} is an extension to the ProGAN architecture, hovers with the ability to control over the disentangled style properties of the generated images. StyleGAN2 \cite{karras2019analyzing} fixed the imperfection of StyleGAN to improve image quality. SNGAN \cite{miyato2018spectral} proposes a novel weight normalization technique called spectral normalization to stabilize the training of the discriminator. It is capable of generating images of better or equal quality relative to the previous training stabilization techniques. MMDGAN \cite{li2017mmd} combines the key ideas in both generative moment matching network (GMMN) and GAN. 

\noindent\emph{Partial image manipulation}.
AttGAN \cite{he2019attgan} applies an attribute classification constraint to the generated image to guarantee the correct change of desired attributes. StarGAN \cite{choi2018stargan} simply uses a single model to perform image-to-image translations for multiple facial properties. STGAN \cite{liu2019stgan} simultaneously improves attribute manipulation accuracy as well as perception quality on the basis of AttGAN. 



\subsection{DeepFake Detection Methods}\label{DeepFake Detection Methods}
Tolosana \etal and Verdoliva \etal \cite{tolosana2020deepfakes,verdoliva2020media} recently make comprehensive surveys on the DeepFake detection methods \cite{wang2019fakespotter,arxiv20_deeprhythm,wang2019cnn,afchar2018mesonet,nguyen2019capsule,nguyen2019multi,yu2019attributing,zhang2019detecting,frank2020leveraging,arxiv20_deepsonar}. Overall, they surveyed thirty-seven papers in total,
all of which are CNN-based and can be classified into three categories depending on their feature inputs: image-based methods, fingerprint-based methods, and spectrum-based methods. Image-based methods directly use images as inputs with various networks to solve the problem. Fingerprint-based methods detect DeepFake with the features of GAN fingerprints. Spectrum-based methods consider that DeepFake artifacts are manifested as replications of spectra in the frequency domain. Thus they propose classifiers based on the spectrum of fake images.

\section{Method}\label{sec:method}
In this section, we give a more detailed discussion on the limitation of GAN-based methods and introduce our post-processing method.

\subsection{Artifact of GAN-Based Image Generation}
For both partial image manipulation \cite{choi2018stargan,he2019attgan,liu2019stgan} and full image synthesis \cite{karras2017progressive,karras2019style,karras2019analyzing}, the generator of GAN-based image generation method has a decoder amplifies the random vectors or feature maps to images. Upsampling is a significant and indispensable design in the decoder. However, it is the upsampling design that makes the GAN-based image generation methods limited. In general, there are three types of upsampling methods: unpooling, transpose convolution and interpolation. It has been studied that transpose convolution results in checkerboard texture \cite{odena2016deconvolution}. In the amplification procedure, unpooling operation assigns zero values to the new pixels. This regular magnification produces special textures that do not exist in real images. For interpolation operation, at an intuitive level, the new pixel values are calculated based on existing pixels, which is regularity. Interpolation brings periodicity into the second derivative signal of images \cite{gallagher2005detection}. Other papers \cite{zhang2019detecting,frank2020leveraging,huang2020fakelocator} also introduced the imperfection of upsampling methods.


\subsection{Shallow Reconstruction to the Rescue}
In this paper, we propose \emph{FakePolisher}, a post-processing method that performs a shallow modification of fake images. Our method is composed of three steps. First, we train a dictionary model with a real image dataset. The learned dictionary forms a subspace that is intrinsically low dimensional, which compactly captures the essential structures and representations of the real images. 
Second, we seek the representation of a DeepFake image using the aforementioned subspace either by linear projection or sparse coding depending on the over-completeness of the learned dictionary. Third, once such a representation is obtained, we reconstruct the `fake-free' version of the DeepFake image by using the said dictionary.\footnote{Throughout the paper, we use `clean' to describe an image that is free of fake patterns. `Clean' and `fake pattern-free' are used interchangeably.} 

Intuitively, we are forcing the DeepFake image to find its `closest' representation, on the subspace that subsequently leads to the reconstructed version of itself free of any fake patterns. By doing so, a shallow reconstruction can already effectively remove the fake patterns while preserving image fidelity to the greatest extent. In this context, deep reconstruction methods, on the contrary, can become futile because they potentially leave traces of upsampling artifacts as discussed above. 
\begin{SCfigure}[]
	\centering 
	\includegraphics[width=0.42\columnwidth]{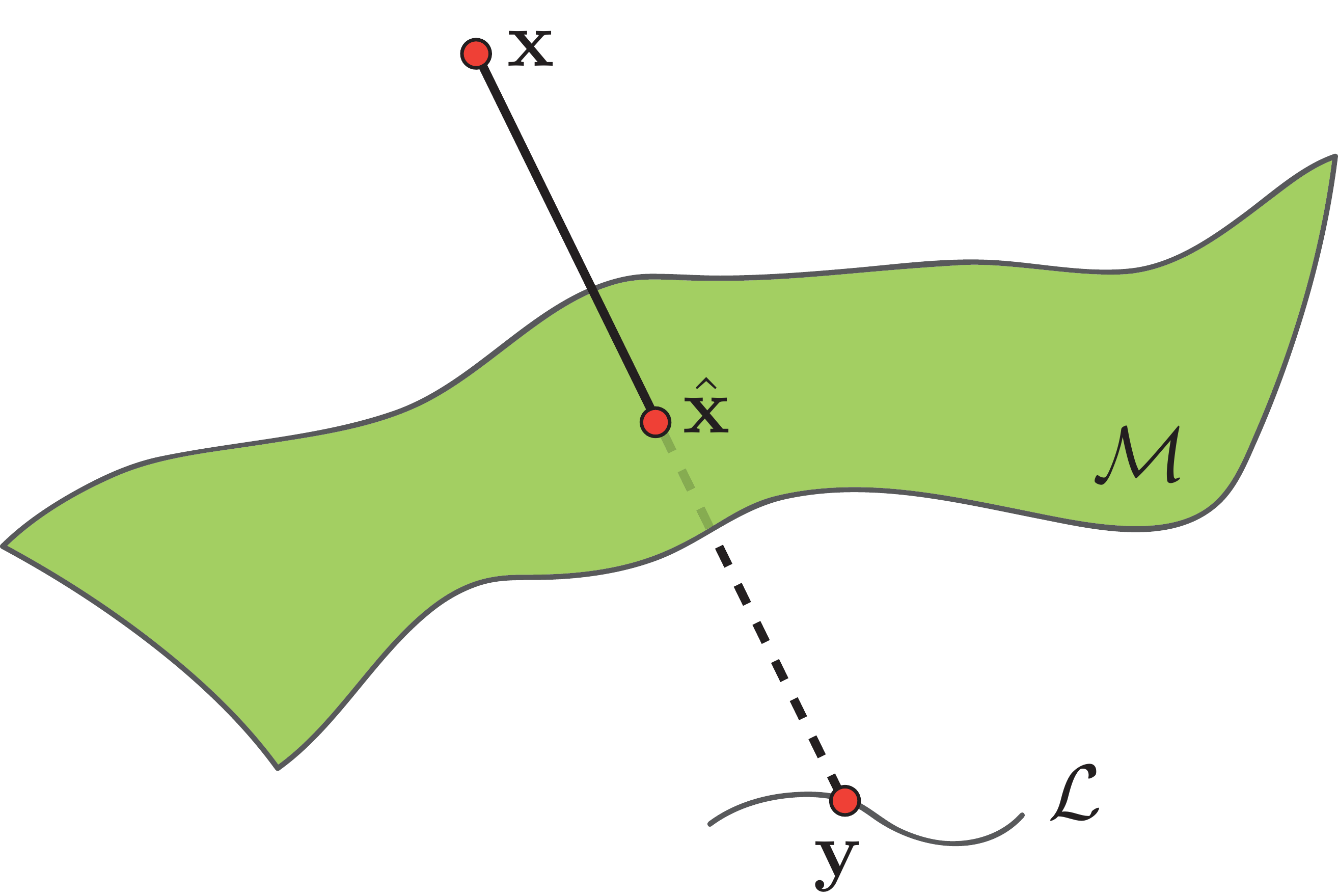}
	\caption{Geometric interpretation of how the shallow reconstruction works in order to bring the DeepFake image $\mathbf{x}$ onto the clean image manifold $\mathcal{M}$ thus finds its `closest' counterpart $\hat{\mathbf{x}}$ on the manifold through the embedded representation $\mathbf{y}$ in the embedding space $\mathcal{L}$.}
	\label{fig:manifold}
\end{SCfigure}

Geometrically, as shown in Figure~\ref{fig:manifold}, the learned dictionary forms an embedding space $\mathcal{L}$ that could be of lower or higher dimensionality. When a DeepFake image $\mathbf{x}$ comes along, we seek the `closest' counterpart $\hat{\mathbf{x}}$ on the clean image manifold $\mathcal{M}$ by first obtaining the embedded representation $\mathbf{y}$, and reconstruct back to the clean image manifold $\mathcal{M}$.


\subsection{Global vs. Local Dictionary Learning}
When learning the dictionary on the real images that are free of fake patterns, one can choose to learn a local patch-based dictionary, leading to a patch-based image reconstruction; or, a global dictionary that spans the entire image, \ie, dictionary atom is of the same size as the image to be reconstructed. Since we can view the global case as a local case with a large patch size, we will mainly discuss patch-based local reconstruction in detail since it already covers both cases. The choice between various patch sizes (local vs. global) is largely dictated by the actual application and the type of images that are being processed. For example, if the images are aligned faces, it is advisable to use a global dictionary since it is more efficient, without the need of patch-by-patch reconstruction. On the other hand, if the images are of ImageNet type, it is more reasonable to use a patch-based dictionary. One note is that in this case, it is still possible to use a global dictionary, it is just we are to foresee a drop in the reconstruction fidelity.

Next, we formulate the patch-based dictionary learning procedure. The training data (patch) matrix $\mathbf{Y}\in \mathds{R}^{d\times n}$ is assumed with dimension $d$. All matrices have their elements arranged column-wise.

Dictionary learning methods have gained much popularity in tackling low-level computer vision problems. One widely adopted such an algorithm is the K-SVD \cite{ksvd}. K-SVD aims to be a natural extension of K-means clustering method with the analogy that the cluster centroids are the elements of the learned dictionary and the cluster memberships are defined by the sparse approximations ($\ell_1$ or $\ell_0$) of the signals in that dictionary. Formally, it provides a solution to the problem:
\begin{align}
\mathrm{(KSVD)}\quad \minimize_{\mathbf{D},\mathbf{X}}\| \mathbf{Y} - \mathbf{D}\mathbf{X} \|_{F}^{2} \quad\mathrm{subject~to}\quad \forall i, \|\mathbf{x}_i\|_0 < K
\end{align}
where $\mathbf{Y}$, $\mathbf{D}$ and $\mathbf{X}$ are the data, the learned dictionary, and the sparse approximation matrix, respectively. Here $\|.\|_0$ is the pseudo-norm measuring sparsity. The sparse approximations of the data elements are allowed to have some maximum sparsity $\|\mathbf{x}\|_0 \leq K$.

In addition to the K-SVD method, we also explore a ubiquitously popular dictionary learning method: principal component analysis (PCA) \cite{pca}. In the original formulation of PCA, it tries to minimize the following objective function in an $\ell_2$ sense. Of course, variants of PCA such as sparse PCA or $\ell_1$-PCA, \etc, can also fit in with ease. Formally, PCA finds a solution to the problem:
\begin{align}
\mathrm{(PCA)}\quad \minimize_{\mathbf{D},\mathbf{X}}\| \mathbf{Y} - \mathbf{D}\mathbf{X} \|_{F}^{2} \quad\mathrm{subject~to}\quad \mathbf{D}^\top\mathbf{D} = \mathbf{I}
\end{align}
where $\mathbf{Y}$, $\mathbf{D}$ and $\mathbf{X}$ are the data, the learned dictionary (principal components), and the dense coefficient matrix, respectively. The regularizer in the PCA optimization ensures that the learned dictionary atoms are orthogonal, which provides maximal reconstruction capability. The learned PCA dictionary is usually overdetermined (or undercomplete), which provides a good complement to the overcomplete K-SVD dictionary.

\subsection{Shallow Reconstruction from Dense vs. Sparse Representation}

Once the said dictionaries are learned from real images that are fake-pattern-free, we can project a DeepFake image patch onto the learned subspace and obtain a new representation on the learned manifold. The dimensionality of such a representation can be lower or higher than the original image-domain representation depending on the overcompleteness of the learned dictionary. 

More specifically, for example, when we want to reconstruct a single image patch $\mathbf{y}\in \mathds{R}^d$ using learned K-SVD dictionary, since it is overcomplete, we resort to pursuit algorithms such as the orthogonal matching pursuit (OMP) \cite{omp} to obtain the sparse coefficient vector $\mathbf{x}$ according to the following optimization:
\begin{align}
\mathrm{(OMP)}\quad \minimize_{\mathbf{x}}\| \mathbf{y} - \mathbf{D}\mathbf{x} \|_{2}^{2} \quad\mathrm{subject~to}\quad \forall i, \|\mathbf{x}\|_0 < \tau
\end{align}

Note that there is a trade-off in choosing the sparsity $\tau$ while using OMP for obtaining the sparse representation. To determine the optimal reconstruction sparsity $\tau$ for the down-stream task, we conduct a pilot experiment that aims at selecting a $\tau$ value that is both relatively small (more efficient for the greedy OMP algorithm) and provides high-quality reconstruction. Also, the sparsity $\tau$ during the OMP step is independent and different from the sparsity $K$ during K-SVD dictionary learning. 

The shallow reconstruction is straight-forward with the learned K-SVD dictionary $\mathbf{D}$ and the obtained sparse representation $\mathbf{x}$. The reconstructed image patch $\hat{\mathbf{y}} = \mathbf{Dx}$. The aforementioned dictionary learning and reconstruction were previously used in various domain-domain mapping problems such as \cite{cvprw14_hallucinate,cvprw15_nir,icip15_sr,icip15_illum,pr16_fkda,btas16_fastfood,bmvc16_invert,pr19_ssr2}.

As a comparison, shallow reconstruction from learned PCA dictionary requires first obtaining a dense representation for the image patch $\mathbf{y}$. As discussed above, the learned PCA  dictionary $\mathbf{D}$ is usually overdetermined, and therefore, the resulting representation vector $\mathbf{x}$ on the manifold will be of lower dimensionality and dense. The representation vector $\mathbf{x}$ can be obtained through a least-square error solution in closed form which is extremely efficient:
\begin{align}
    \mathbf{x} = (  \mathbf{D}^\top  \mathbf{D} )^{-1} \mathbf{D}^\top \mathbf{y}
\end{align}

To further make the shallow reconstruction using PCA more versatile, one can control what dimensions contribute more during the reconstruction by involving a selector vector $\mathbf{s}\in \mathds{R}^d$ that embeds \eg, prior knowledge such as confidence or importance of each dimension. In this case, the representation vector $\mathbf{x}$ can be obtained by incorporating a diagonal selector matrix $\mathbf{S}$, where $\mathbf{S} = \mathrm{diag}(\mathbf{s})$. The solution becomes:
\begin{align}
  \hat{\mathbf{x}} &= [  \mathbf{(SD)}^\top  \mathbf{(SD)} ]^{-1} \mathbf{(SD)}^\top \mathbf{Sy} = [ \mathbf{D}^\top \mathbf{S}^\top \mathbf{S} \mathbf{D} ]^{-1}  \mathbf{D}^\top \mathbf{S}^\top  \mathbf{S}\mathbf{y} \\
  &= [ \mathbf{D}^\top (\mathbf{S}^\top \mathbf{S}) \mathbf{D} ]^{-1}  \mathbf{D}^\top (\mathbf{S}^\top \mathbf{S})  \mathbf{y}
\end{align}

It can be observed that when $\mathbf{S}^{\top}\mathbf{S}$ is close to the identity matrix $\mathbf{I}$, the $\hat{\mathbf{x}}$ is close to the original $\mathbf{x}$. The same principal can be applied to the aforementioned K-SVD sparse reconstruction. Also, the selector vector $\mathbf{s}$ can be both real-numbered (dimension re-weighting) or binary (dimension selection). 

The shallow reconstruction is also straight-forward with the learned PCA dictionary $\mathbf{D}$ and the obtained dense representation $\mathbf{x}$, with the reconstructed image patch $\hat{\mathbf{y}} = \mathbf{Dx}$.

Both K-SVD and PCA reconstructions are shallow in the sense that they can bring back the manifold representations to the image domain with a single-step projection. More importantly, with the reconstruction being shallow and single-step, it does not induce unnecessary fake patterns, as commonly found in those DeepFake images produced or manipulated by deep generative models.



\subsubsection{Discussion: Comparison with Denoising Autoencoder and DefenseGAN} 

Denoising Autoencoder (DAE) \cite{dae} is an early attempt for image deep reconstruction, especially for the image denoising task. Compared to the shallow reconstruction we have discussed in this work, DAE is usually comprised of several layers of fully connected or convolutional layers in both the encoder and the decoder, interlaced with non-linearity.

Apart from being much deeper and non-linear, the training process of the DAE takes the noisy version of the data as input, and the reconstructed version is then compared with the clean version, whose discrepancy amounts to the loss that needs to be minimized by tuning the weights in the encoder and the decoder. The model usually works well when the input image is corrupted with the same noise that the model has seen during the training process. 

As a comparison, our shallow reconstruction method has the following main advantages: (1) the model is shallow and linear, which is much easier to train with little to none tuning required, and it is much more data efficient; (2) the training only requires a clean version of the images (as compared to the clean and noisy pairs as in the DAE), and such reconstruction can deal with arbitrary non-clean versions of the data, be it some types of noises, or DeepFake patterns that need to be removed.  

Another line of recent work in the context of removing adversarial noise through deep image reconstruction is the DefenseGAN method \cite{samangouei2018defense}. The idea is to train a GAN generator based only on clean images and any adversarially noisy image can be noise-removed by DefenseGAN deep reconstruction. In some sense, it seems that it can be re-purposed for reconstructing DeepFake images that are free of fake patterns. However, a major issue remains because the deep reconstruction in the DefenseGAN also results in fake patterns, which is something our proposed shallow reconstruction is trying hard to prevent.

\section{Experiments}\label{sec:exp}
To demonstrate the effectiveness of our shallow reconstruction, we propose two different validation methods. One is to test the reconstructed images on various fake image detection methods, which indicates whether there exists some relation between these detection methods and the artifacts (\ie, manipulation footprint). The other is using metrics to measure the similarity between fake images and reconstructed images, quantitatively measure our reconstructed change magnitude.
These two validation methods are able to confirm the usefulness of our method. We also give some concrete examples to show the fake images and our reconstructed ones.

\begin{SCfigure}[]
	\centering 
	\includegraphics[width=0.55\columnwidth]{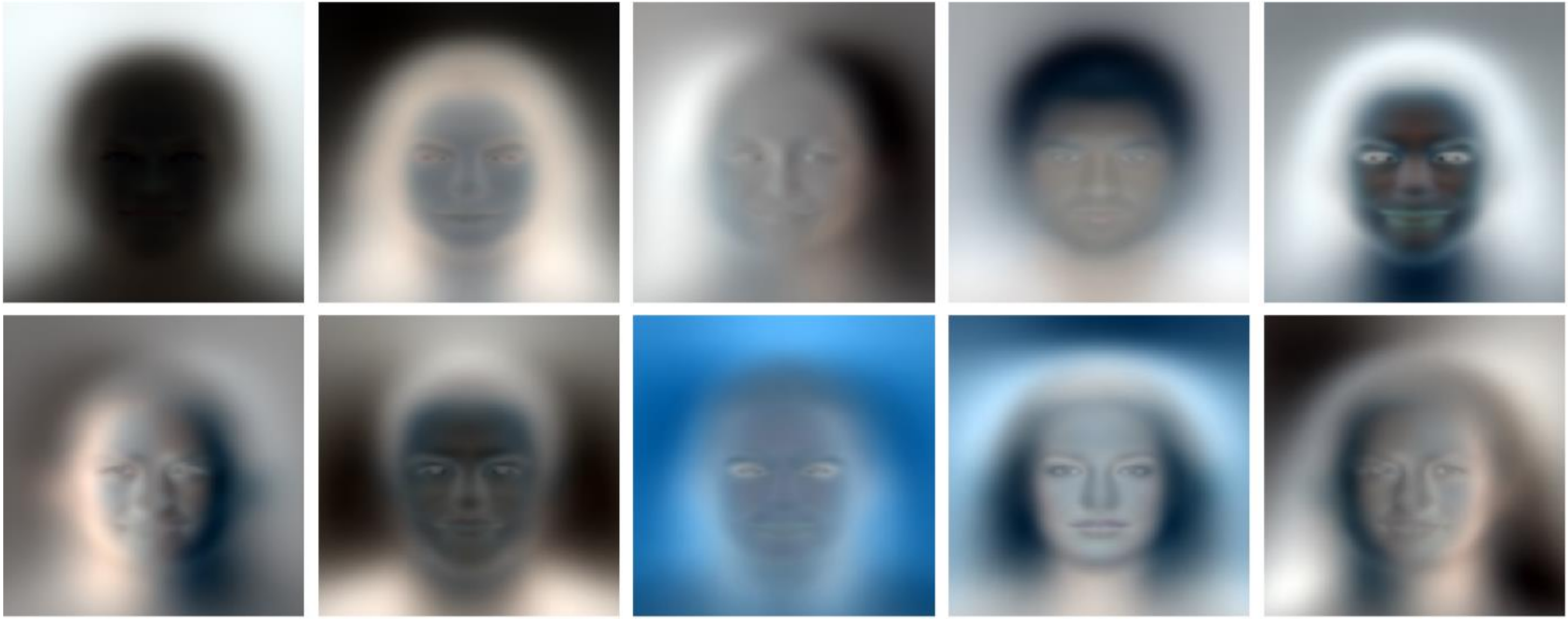}
	\caption{The PCA dictionary generated by us has 10,000 components. The size of each component is 224*224*3. Here we show the images of the first ten principal components.}
	\label{fig:PCA_component}
\end{SCfigure}

\subsection{Experimental Setup}
\textbf{Subject Detection Methods and Dataset:}
We choose three state-of-the-art fake detection methods to verify the validity of our method, \ie, \textbf{GANFingerprint} (fingerprint-based method) \cite{yu2019attributing}, \textbf{CNNDetector} (image-based method) \cite{wang2019cnn}, and \textbf{DCTA} (spectrum-based method) \cite{frank2020leveraging}. 
We use CelebA \cite{liu2018large},  LSUN \cite{yu2015lsun}, and FFHQ \cite{karras2019style} as the real image dataset. 
CelebA and FFHQ are the human face dataset while LSUN includes the images of different rooms such as classroom, bedroom, \etc, which are widely used in previous work. 
Then, we leverage a total of 16 GAN-based methods for fake image generation on these datasets. In particular, 
ProGAN \cite{karras2017progressive}, SNGAN \cite{miyato2018spectral}, CramerGAN \cite{bellemare2017cramer} and MMDGAN \cite{li2017mmd} are the GAN-based image generation methods used by \textbf{GANFingerprint} and \textbf{DCTA}. 
For each GAN-based image generation method, the size of the testing dataset is 10,000. 
In \textbf{CNNDetector}, they choose 13 GAN-based image generation methods as the testing dataset. The methods include ProGAN \cite{karras2017progressive}, StyleGAN \cite{karras2019style} , BigGAN \cite{brock2018large}, CycleGAN \cite{zhu2017unpaired}, StarGAN \cite{choi2018stargan}, GauGAN \cite{park2019gaugan}, CRN \cite{shi2016end}, IMLE \cite{li2018implicit}, SITD \cite{chen2018learning}, SAN \cite{dai2019second}, DeepFakes \cite{rossler2019faceforensics++}, StyleGAN2 \cite{karras2019analyzing}, and Whichfaceisreal \cite{whichfaceisreal}. The size of the testing dataset of these GAN-based image generation methods range from hundreds to thousands. The objects in the datasets of CycleGAN, ProGAN, StyleGAN and StyleGAN2 have two or more categories. For example, in StyleGAN, it has three different categories: bedroom, car, cat. 
The datasets of other GANs have only one category. 

\begin{SCfigure}[]
	\centering 
	\includegraphics[width=0.55\columnwidth]{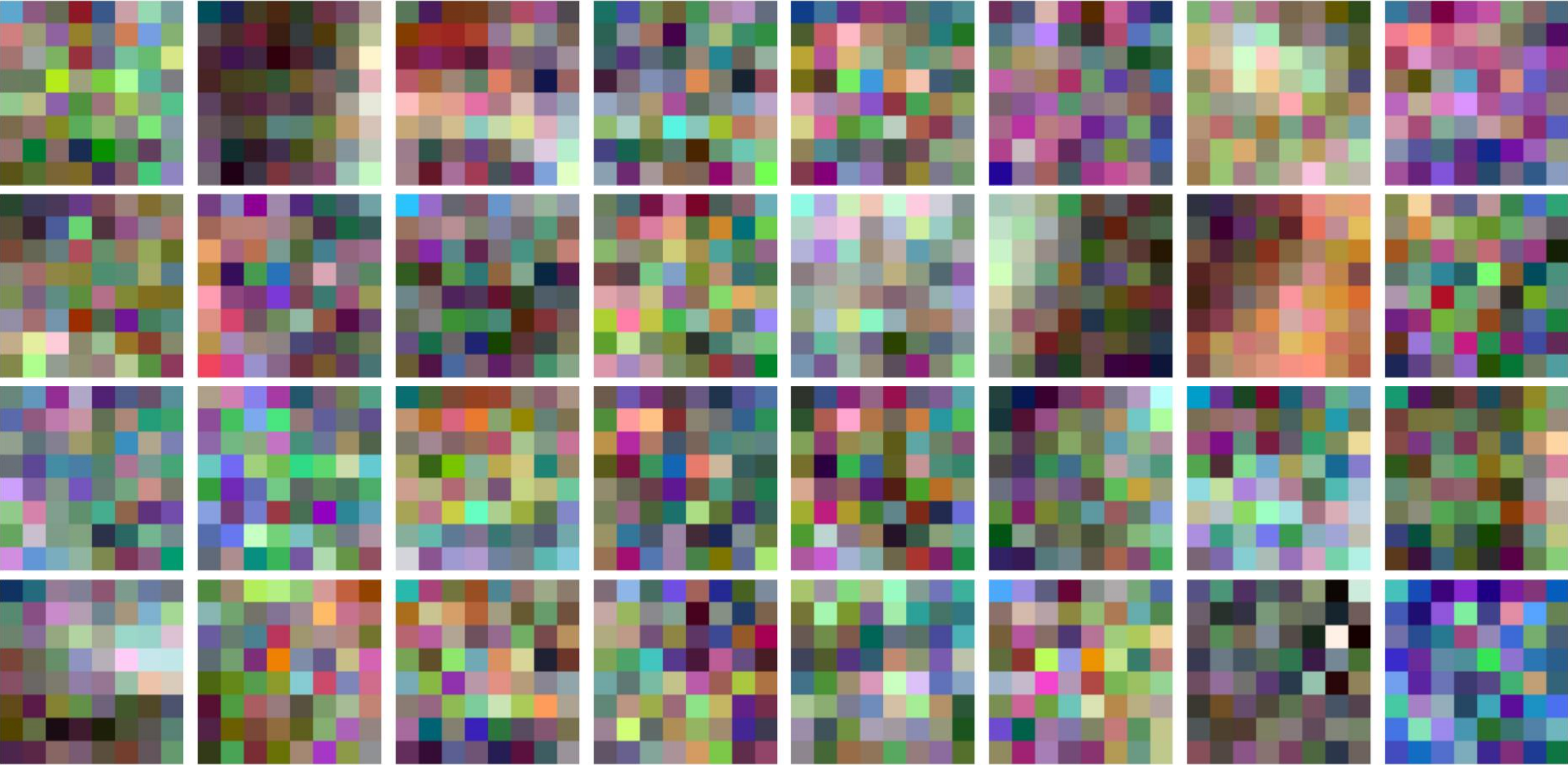}
	\caption{The K-SVD dictionary generated by us has 5,000 components. The size of each component is 8*8*3. The components of K-SVD dictionary are not sequenced. Thus we randomly show images of 32 components.}
	\label{fig:KSVD_component}
\end{SCfigure}

\textbf{Evaluation Settings:}
For PCA reconstruction, we use 50,000 real human images of CelebA to train the PCA dictionary model. The component number of PCA dictionary model is 10,000. For K-SVD reconstruction, we use 100,000 patches to train a K-SVD model of 5,000 components. Each patch is of size 8*8, clipped from real images of CelebA. The number of nonzero coefficients in the training procedure is 15. In K-SVD reconstruction, we drop 10\% pixels of the fake image before reconstruction. This is an important procedure in K-SVD reconstruction for that it can destroy the fake textures of the fake images. What's more, it needs a lot of time to produce one K-SVD image. Therefore, we choose 200 of the 5,000 components of the K-SVD dictionary to reconstruct fake images. The reconstructed images of using 200 or 5,000 components are similar while the reconstruction time is significantly reduced. In the reconstructing procedure, the number of nonzero coefficients is 20. The graphical representation of the PCA dictionary and K-SVD dictionary are shown in Figure \ref{fig:PCA_component} and Figure \ref{fig:KSVD_component}.

\textbf{Metrics:}
The main metric is the detection accuracy of the methods. We compare the detection accuracy of fake images and reconstructed images for each method. In addition, we also use cosine similarity (COSS), peak signal-to-noise ratio (PSNR) and structural similarity (SSIM) for measuring the similarity between fake image and its corresponding reconstructed image. COSS is a common similarity metric that measures the cosine of the angle. We transform the RGB images to vectors before calculating COSS. PSNR is the most commonly used measurement for the reconstruction quality of lossy compression. SSIM is one of the most popular and useful metrics for measuring the similarity between two images. COSS, PSNR and SSIM metrics are better if a higher value is provided. The value ranges of COSS and SSIM are both in [0,1].

All the experiments were run on a Ubuntu 16.04 system with an Intel(R) Xeon(R) CPU E5-2699 with 196 GB of RAM, equipped with four Tesla V100 GPU of 32G RAM.

\begin{figure}[tbp]
	\centering 
	\includegraphics[width=0.7\columnwidth]{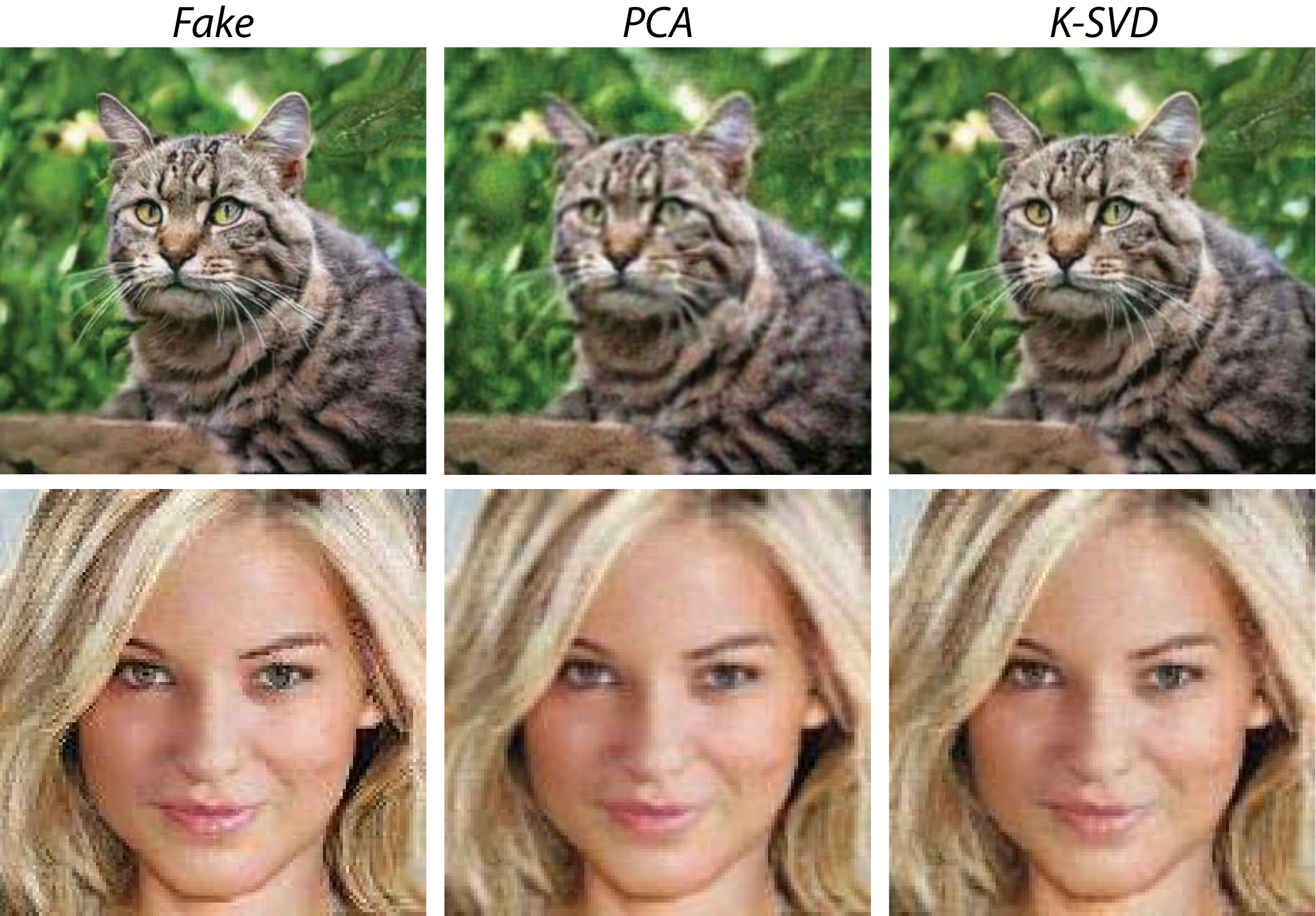}
	\caption{In the first row, the images in turn are the fake image produced by StyleGAN \cite{karras2019style}, PCA reconstructed image and K-SVD reconstructed image. In the second row, the images in turn are the fake image produced by SNGAN \cite{miyato2018spectral}, PCA reconstructed image and K-SVD reconstructed image.}
	\label{fig:reconstruction_person}
\end{figure}


\begin{figure}[tbp]
	\centering 
	\includegraphics[width=0.9\columnwidth]{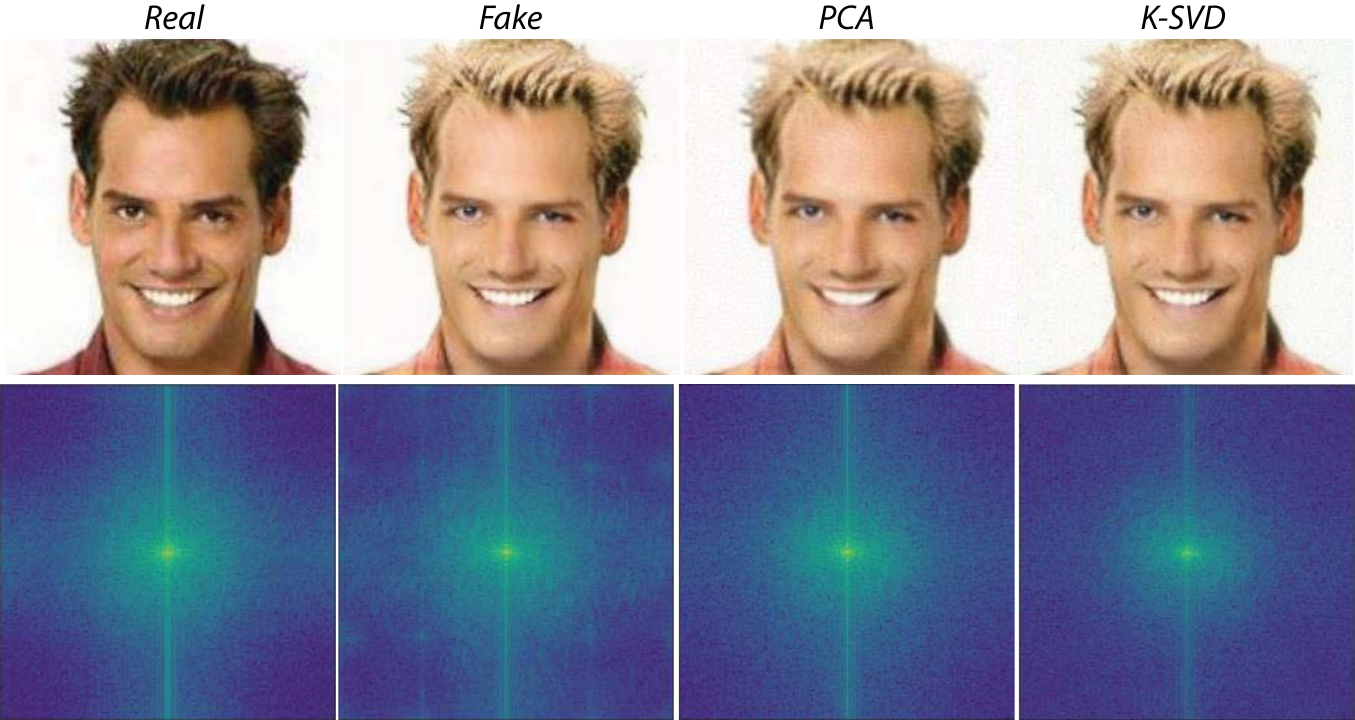}
	\caption{In the first row, the images in turn are a real image from CelebA, a fake image produced by StarGAN \cite{choi2018stargan}, PCA reconstructed image and K-SVD reconstructed image. In the second row, the images are the spectrum corresponding to the images above. As \cite{zhang2019detecting} mentioned, the GAN-synthesised images have obvious artifacts in the frequency spectrum. In the spectrum of the fake image, there are bright blobs at 1/4 and 3/4 of the width/height. In the spectrum of reconstructed images, the artifact does not exist. This means that our method effectively reduces fake texture from the fake image.}
	\label{fig:spectrum}
\end{figure}

\subsection{Examples of Reconstructed Image}
Figure \ref{fig:reconstruction_person} gives the reconstructed image examples of our method, on a cat and a human, respectively. In the second row, PCA and K-SVD reconstruct the fake image successfully. In the first row, we can see that the fidelity of PCA reconstructed image is not as good as K-SVD reconstructed image. The reason is that the PCA dictionary we used is trained by real images of humans instead of cats.
Thus, this suggests using K-SVD to reconstruct fake images if PCA dictionary with the same category is not available.

To analyze whether the reconstructed images contain artifacts (\ie, manipulation footprints), we analyze the spectrums of the real image, fake image, PCA reconstructed image, K-SVD reconstructed image (see Figure \ref{fig:spectrum}). We can observe that only the spectrum of the fake image has bright blobs at 1/4 and 3/4 of the width/height. The blobs correspond to the manipulation footprints in the fake image. We also verified that the PCA and K-SVD reconstruction methods can reduce artifacts on other types of images such as cat, bedroom, \etc.

\begin{table*}[tbp]
\footnotesize
\centering
\caption{Detection accuracy before \& after reconstruction of GAN-synthesized images in GANFingerprint}
\setlength{\tabcolsep}{3.5pt}
\begin{tabular}{c|c c c c c|c c c c c}
\toprule

\multirow{2}{*}{Accuracy(\%)} &  \multicolumn{5}{c|}{ProGAN (Pro)} & \multicolumn{5}{c}{SNGAN (SN)}   \\ 
                  & CelebA  & Pro  & SN &Cramer & MMD & CelebA  & Pro  & SN & Cramer & MMD \\ \midrule
Fake                & 0.03  & 99.91  & 0.01 & 0.03  &0.02 & 0.07  & 0.01  & 99.75 & 0.05 & 0.12  \\ \hline
PCA-reconstructed   & 88.90 \textcolor{blue}{(+88.87)}  & 6.99   \textcolor{red}{(-92.92)}& 0.21 \textcolor{blue}{(+0.20)}& 0.07 \textcolor{blue}{(+0.04)}& 3.83 \textcolor{blue}{(+3.81)}& 46.10  \textcolor{blue}{(+46.03)}& 0.12  \textcolor{blue}{(+0.11)}& 50.86 \textcolor{red}{(-48.89)}& 0.16 \textcolor{blue}{(+0.11)}& 2.76 \textcolor{blue}{(+2.64)}\\ \hline
K-SVD-reconstructed  &  21.50 \textcolor{blue}{(+21.47)}& 78.10  \textcolor{red}{(-21.81)}&0.10  \textcolor{blue}{(+0.09)}&0.20  \textcolor{blue}{(+0.17)}&0.10  \textcolor{blue}{(+0.08)}& 48.15  \textcolor{blue}{(+48.08)}& 4.60  \textcolor{blue}{(+4.59)}& 46.35 \textcolor{red}{(-53.40)}& 0.60 \textcolor{blue}{(+0.55)}& 0.30 \textcolor{blue}{(+0.18)}\\ \midrule

\multirow{2}{*}{Accuracy(\%)} &  \multicolumn{5}{c|}{CramerGAN (Cramer)} & \multicolumn{5}{c}{MMDGAN (MMD)}    \\                     & CelebA  & Pro  & SN &Cramer & MMD & CelebA  & Pro  & SN & Cramer & MMD \\ \midrule

Fake                & 0.00  & 0.02  & 0.02 & 99.76 &  0.20 & 0.11  & 0.01  & 0.04 & 0.27 & 99.57 \\ \hline
PCA-reconstructed   & 54.85  \textcolor{blue}{(+54.85)}& 0.35 \textcolor{blue}{(+0.33)}& 0.93 \textcolor{blue}{(+0.91)}& 35.07 \textcolor{red}{(-64.69)}& 8.80 \textcolor{blue}{(+8.60)}& 45.94  \textcolor{blue}{(+45.83)}& 0.13  \textcolor{blue}{(+0.12)}& 0.20 \textcolor{blue}{(+0.16)}& 0.03 \textcolor{red}{(-0.24)}& 53.70 \textcolor{red}{(-45.87)}\\ \hline
K-SVD-reconstructed  &  28.70 \textcolor{blue}{(+28.70)}& 14.90 \textcolor{blue}{(+14.88)} & 0.10 \textcolor{blue}{(+0.08)}& 55.60 \textcolor{red}{(-44.16)}& 0.70 \textcolor{blue}{(+0.50)}&  47.40 \textcolor{blue}{(+47.29)}& 14.20   \textcolor{blue}{(+14.19)}& 0.30 \textcolor{blue}{(+0.26)}& 0.70 \textcolor{blue}{(+0.43)}& 37.40 \textcolor{red}{(-62.17)}\\

\bottomrule
\end{tabular}
\label{Table:GANFingerprint_Acc}
\end{table*}
\begin{table}[tbp]
\footnotesize
\centering
\caption{Similarity between fake image \& reconstructed image of GANs in GANFingerprint \& DCTA}
\setlength{\tabcolsep}{3.5pt}
\begin{tabular}{c c c c c c}
\toprule

\multicolumn{2}{c}{} &  ProGAN & SNGAN & CramerGAN & MMDGAN\\ 
\toprule
\multirow{3}{*}{PCA} &  COSS   & 0.999 &0.999 &0.998 &0.999   \\ 
&PSNR   & 32.33 & 32.67 & 31.85 & 32.28   \\ 
&SSIM  & 0.960 &0.960 &0.957 & 0.959 \\ \hline
\multirow{3}{*}{K-SVD} &  COSS   & 0.999 & 0.999 & 0.999 & 0.999   \\
&PSNR   & 33.224  & 33.526  & 32.897 & 33.304   \\ 
&SSIM  & 0.972 & 0.972 & 0.971 & 0.972
\\ 

\bottomrule
\end{tabular}
\label{Table:GANFingerprint_Similarity}
\end{table}

\subsection{GANFingerprint}
In the original experiment \cite{yu2019attributing} of \textbf{GANFingerprint}, their method can successfully detect whether an input image is real or fake with high accuracy. It can even judge which GAN-based image generation method is used to produce the fake image. In our experiment, we randomly choose 10,000 real images from CelebA. Then, 
for each GAN-based image generation method (\ie, ProGAN, SNGAN, CramerGAN, MMDGAN), we produce 10,000 fake images, resulting in a total of 40,000. For each of PCA and K-SVD reconstruction method, we produce 40,000 images from these 40,000 fake images. 

Table \ref{Table:GANFingerprint_Acc} summarizes the detailed detection accuracy of \textbf{GANFingerprint}. We use ProGAN as an example to explain the data in Table \ref{Table:GANFingerprint_Acc} (\ie, columns 2-6, rows 2-4).
For \textbf{ProGAN (Pro)}, it has five sub-items (columns): CelebA, ProGAN (Pro), SNGAN (SN), CramerGAN (Cramer), MMDGAN (MMD). They represent the possibility that the input ProGAN fake images be considered as one of them. \textbf{Fake}, \textbf{PCA-reconstructed} and \textbf{K-SVD-reconstructed} represent the type of input images. 
The row of \textbf{Fake} shows the results with fake images as inputs. As we can see, \textbf{GANFingerprint} can accurately classify the 10,000 images generated by ProGAN into subitem ProGAN with a detection accuracy of 99.91\%. 
In the table, we highlight the difference in detection accuracy between reconstructed images and fake images (\eg, by color and number). For example, in the row \textbf{PCA-reconstructed}, when we put the 10,000 PCA-reconstructed images into \textbf{GANFingerprint}, it misclassifies most of the images into CelebA (\ie, real images). The ratio of images classified into CelebA raises from 0.03\% to 88.90\%. We use blue color and \textcolor{blue}{(+88.87)} to highlight the difference. Similarly, the ratio of images classified into Pro decreases from 99.91\% to 6.99\%. We use red color and \textcolor{red}{(-92.92)} to show the difference. We can see that most of the fake images generated by ProGAN are misclassifies to be real images after our shallow reconstruction.

Similar conclusions could also be reached for the other three GAN-based image generation methods. PCA reconstruction reduces all of their classification accuracy effectively. K-SVD reconstruction also reduces the accuracy. The attack by PCA reconstruction shows slightly higher results compared with K-SVD reconstruction.

Table \ref{Table:GANFingerprint_Similarity} shows the similarity between the fake images and reconstructed images. For both PCA and K-SVD, we use COSS, PSNR and SSIM as the metrics for similarity measurement. In the column of \textbf{ProGAN}, we can see that the values of COSS and SSIM are near 1.0, and the value of PSNR is more than 30, indicating high similarity. 
Likewise, for the other three GAN-based image generation methods, the reconstructed images are also very similar to the fake image counterparts.
Compared with PCA reconstruction, images by K-SVD show higher similarity to original fake images.

\begin{table}[tbp]
\footnotesize
\centering
\caption{Detection accuracy before \& after reconstruction of GAN-synthesized images in DCTA}
\setlength{\tabcolsep}{3.5pt}
\begin{tabular}{c c}
\toprule

{} &  Accuracy(\%) \\ 
\toprule

Fake                & 88.99   \\ 
PCA-reconstructed   & 16.42 \textcolor{red}{(-72.57)}  \\ 
K-SVD-reconstructed  & 20.44 \textcolor{red}{(-68.55)}  \\ 

\bottomrule
\end{tabular}
\label{Table:DCTA_Acc}
\end{table}

\begin{table*}[tbp]
\scriptsize
\centering
\caption{Detection accuracy before \& after reconstruction of GAN-synthesized images in \textbf{CNNDetecion}}
\setlength{\tabcolsep}{2pt}
\begin{tabular}{c c c c c c c c c c c c c c c}
\toprule

 \multicolumn{2}{c}{Accuracy(\%)/AP}& ProGAN & StyleGAN & BigGAN & CycleGAN & StarGAN & GauGAN & CRN & IMLE & SITD & SAN & DeepFakes & StyleGAN2 & Whichfaceisreal \\ 
 \toprule
\multirow{4}{*}{\rotatebox{90}{prob0.1}} & Real \& Fake  & 99.9/99.9 & 87.1/99.6 & 70.2/84.5 & 85.2/93.5 & 91.7/98.2 & 78.9/89.5 & 86.3/98.2 & 86.2/98.4 & 90.3/97.2 & 50.5/70.5 & 53.5/89.0 & 84.4/99.1 & 83.6/93.2   \\ 
& Real    & 100/- & 99.9/- & 93.5/- & 91.5/- & 96.7/- & 93.0/- & 72.7/-  & 72.7/- & 93.9/-  & 99.1/-  & 99.9/- & 99.9/- & 92.9/- \\ 
& Fake   & 99.9/- & 74.2/- & 46.8/- & 78.8/- & 86.7/- & 64.8/- & 99.8/- & 99.8/- & 86.7/- & 1.83/- & 6.86/- & 68.8/- & 74.3/- \\ 
  &  PCA   & 42.3 \textcolor{red}{(-57.6)}/- & 3.90 \textcolor{red}{(-70.3)}/- & 12.3 \textcolor{red}{(-34.5)}/- & 35.8 \textcolor{red}{(-43.0)}/- & 36.0 \textcolor{red}{(-50.7)}/- & 14.2 \textcolor{red}{(-50.6)}/- & 6.50 \textcolor{red}{(-93.3)}/-  & 19.4 \textcolor{red}{(-80.4)}/- & 3.89 \textcolor{red}{(-82.8)}/- & 3.20 \textcolor{blue}{(+1.37)}/- & 1.33 \textcolor{red}{(-5.53)}/- & 11.9 \textcolor{red}{(-56.9)}/- & 1.40 \textcolor{red}{(-72.9)}/-  \\ 
  &  K-SVD  & 94.9 \textcolor{red}{(-5.0)}/- & 33.7 \textcolor{red}{(-40.5)}/- & 30.0 \textcolor{red}{(-16.8)}/- & 68.7 \textcolor{red}{(-10.1)}/- & 48.0 \textcolor{red}{(-38.7)}/- & 51.0 \textcolor{red}{(-13.8)}/- & 79.0 \textcolor{red}{(-20.8)}/- & 88.0 \textcolor{red}{(-11.8)}/- & 45.0 \textcolor{red}{(-41.7)}/- & 8.0 \textcolor{blue}{(+6.17)}/- & 0.00 \textcolor{red}{(-6.86)}/- & 33.5 \textcolor{red}{(-35.3)}/- & 50.0 \textcolor{red}{(-24.3)}/- \\ \hline
 \multirow{4}{*}{\rotatebox{90}{prob0.5}} & Real \& Fake & 100/100 & 73.4/98.5 & 59.0/88.2 & 80.8/96.8 & 81.0/95.4 & 79.3/98.1 & 87.6/98.9 & 94.1/99.5 &78.3/92.7 & 50.0/63.9 & 51.1/66.3 & 68.4/98.0 & 63.9/88.8   \\ 
  & Real    & 100/- & 99.9/- & 99.1/-  & 98.6/-  & 99.3/-  & 99.4/- & 99.2/-  &99.2/- & 92.8/-  & 100/- & 99.4/- & 99.9/-  & 99.2/- \\ 
  & Fake   & 100/- & 46.9/- & 18.9/- & 62.9/- & 62.7/- & 59.2/- & 76.0/- & 88.9/- & 63.9/- & 0.00/- & 2.5/- & 36.9/- & 28.6/-   \\ 
 &  PCA   &  71.6 \textcolor{red}{(-28.4)}/- & 3.00 \textcolor{red}{(-43.9)}/- & 6.45 \textcolor{red}{(-12.5)}/- & 30.9 \textcolor{red}{(-32.0)}/- & 42.1 \textcolor{red}{(-20.6)}/- & 22.8 \textcolor{red}{(-36.4)}/- & 4.36 \textcolor{red}{(-71.6)}/- & 16.7 \textcolor{red}{(-72.2)}/- & 1.12 \textcolor{red}{(-62.8)}/- & 0.00 \textcolor{red}{(0)}/- & 1.89 \textcolor{red}{(-0.61)}/- & 6.84 \textcolor{red}{(-30.1)}/- & 0.70 \textcolor{red}{(-27.9)}/- \\ 
  &  K-SVD  & 96.7 \textcolor{red}{(-3.30)}/- & 20.7 \textcolor{red}{(-26.2)}/- & 9.00 \textcolor{red}{(-9.90)}/- & 44.2 \textcolor{red}{(-18.7)}/- & 37.0 \textcolor{red}{(-25.7)}/- & 44.0 \textcolor{red}{(-15.2)}/- & 22.0 \textcolor{red}{(-54.0)}/- & 60.0 \textcolor{red}{(-28.9)}/- & 36.0 \textcolor{red}{(-27.9)}/- & 0.00 \textcolor{red}{(0)}/- & 2.00 \textcolor{red}{(-0.50)}/- & 13.0 \textcolor{red}{(-23.9)}/- & 18.0 \textcolor{red}{(-10.6)}/- \\ 

\bottomrule
\end{tabular}
\label{Table:CNNDetector_Acc}
\end{table*}

\subsection{DCTA}
As mentioned above, \textbf{DCTA} has a same testing dataset as \textbf{GANFingerprint}. In the original experiment of \textbf{DCTA}, it transformed the images into spectrum images before classification. 
We follow the exact same evaluation setting, except that we use the reconstructed images to replace the fake images. Overall,
the number of testing images in our experiment is 48,000. Each category of CelebA, ProGAN, SNGAN, CramerGAN, MMDGAN, has 9,600 images. The PCA-reconstructed and K-SVD-reconstructed images used are the same as that in \textbf{GANFingerprint}. The number of reconstructed images is also 48,000. 
Table \ref{Table:DCTA_Acc} summarizes detection accuracy decrease.
When using PCA-reconstruction, the accuracy decreases from 88.99\% to 16.42\%. Similarly, 20.44\% accuracy decreases when using K-SVD.
The experimental results demonstrate the effectiveness of reconstructed images in misleading the fake detectors.

\begin{table*}[tbp]
\scriptsize
\centering
\caption{Similarity between fake image \&  reconstructed image of GANs in \textbf{CNNDetection}}
\setlength{\tabcolsep}{2pt}
\begin{tabular}{c c c c c c c c c c c c c c c c c c c c c c c}
\toprule

\multicolumn{2}{c}{\multirow{2}{*}{}}  & \cellcolor{gray!40} BigGan   & DeepFakes  & \cellcolor{gray!40} GauGAN  &IMLE  & \cellcolor{gray!40} SAN  & SITD  & \cellcolor{gray!40} StarGAN  & Whichfaceisreal  &  \multicolumn{6}{>{\columncolor{gray!40}}c}{CycleGAN} & \multicolumn{3}{c}{StyleGAN} &  \multicolumn{4}{>{\columncolor{gray!40}}c} {StyleGAN2}\\


                  \multicolumn{2}{c}{} & -  & person  & - & road & - & -  & person  & person  & horse & zebra & winter  & orange  & apple & summer & bedroom & car  & cat &horse & church & car & cat \\ 
                  
                  \toprule
\multirow{3}{*}{PCA} &  COSS               & 0.996  & 0.999  & 0.996 & 0.999  & 0.989 & 0.987  & 0.999  & 0.997 & 0.997 & 0.995 & 0.996  & 0.998  & 0.997 & 0.995 & 0.998 & 0.994  & 0.998  & 0.997  & 0.997 & 0.995 & 0.999 \\ 
&PSNR               & 29.14  & 43.94  & 29.62 & 32.72  & 25.29 & 29.29  & 37.08  & 29.21 & 29.28 & 27.39 & 28.50  & 31.51  & 30.19 & 28.10 &  30.53 & 25.98  & 32.47  & 29.91  & 28.07 & 27.22 & 34.70\\ 
&SSIM               & 0.897  & 0.993  & 0.902 & 0.945  & 0.821 & 0.886  & 0.975  & 0.899 & 0.896 & 0.870 & 0.885  & 0.917  & 0.910 & 0.877 &  0.916 & 0.844  & 0.933  & 0.899  & 0.881 & 0.864 & 0.954 \\ \hline
\multirow{3}{*}{K-SVD} &  COSS   & 0.998  & 0.999  & 0.999 & 0.999  & 0.999 &0.991  & 0.999  & 0.999 & 0.999 & 0.999 & 0.999 & 0.999 & 0.999 & 0.999 &
  0.999 & 0.999  & 0.999  & 0.999  & 0.998 & 0.999 & 0.999\\  
&PSNR   & 32.17  & 39.46  & 39.14 & 39.58  &36.51 & 31.38  & 38.63  & 37.50 & 33.55 &32.39 &32.58 &33.70 &33.48 &31.99 &  33.56 & 33.50  & 34.93  & 34.24  & 31.18 & 34.34 & 37.35\\  
&SSIM   & 0.961  & 0.988  & 0.965 & 0.986  & 0.986 & 0.962  & 0.987  & 0.980 &0.969 &0.967 &0.966 &0.961 &0.966 &0.961&  0.968 & 0.971  & 0.973  & 0.969 & 0.956 & 0.973 & 0.980 \\ 
\toprule

\multicolumn{2}{c}{\multirow{2}{*}{}}  &  \multicolumn{20}{>{\columncolor{gray!40}}c}{ProGAN} & CRN\\ 
                  \multicolumn{2}{c}{} &  airplane & motorbike  & tvmonitor  & horse & sofa  & car  &  pottedplant  & diningtable & sheep  & bottle & person  & train  & dog  & cow & bicycle  & cat & bird  & boat & chair & bus & road \\
                  
                \toprule
                  
\multirow{3}{*}{PCA} &  COSS               & 0.998  & 0.995  & 0.997 & 0.996  & 0.998 & 0.995  & 0.994  & 0.996 & 0.996 & 0.997 & 0.996  & 0.996  & 0.997 & 0.997 &  0.994 & 0.997  & 0.997  & 0.996  & 0.997 & 0.995 & 0.998\\ 
&PSNR               & 29.74  & 26.14  & 28.17 & 27.94  & 29.61 & 27.49  & 26.15  & 27.12 & 28.13 & 29.21 & 28.95  & 27.49  & 29.59 & 28.24 &  26.02 & 30.39  & 29.16  & 27.94  & 28.74 & 26.68 & 31.13\\ 
&SSIM               & 0.913  & 0.867  & 0.898 & 0.883  & 0.991 & 0.884  & 0.858  & 0.881 & 0.878 & 0.908 & 0.901  & 0.875  & 0.905 & 0.882 &  0.860 & 0.917  & 0.899  & 0.880  & 0.904 & 0.867 & 0.932 \\ \hline
\multirow{3}{*}{K-SVD} &  COSS   & 0.999  & 0.998  & 0.998  &0.999 & 0.999 & 0.998  & 0.998  & 0.998 & 0.998 & 0.998 & 0.998  & 0.998  & 0.999 & 0.998 &  0.998 & 0.999  & 0.998  & 0.998  & 0.999 & 0.998 & 0.999\\  
&PSNR   & 33.73  & 30.76  & 32.37 & 32.49 & 32.94  & 32.01  & 30.43 & 31.73 & 32.01 & 32.49  & 33.13  & 31.65 & 33.34 &  32.00 & 30.70  & 33.90  & 32.82  &31.88 & 32.78 & 31.37 & 37.49\\  
&SSIM   & 0.974  & 0.965  & 0.972 & 0.968  & 0.970 & 0.970  & 0.959 & 0.968 & 0.963 & 0.968 & 0.973  & 0.963  & 0.970 & 0.965 & 0.963 & 0.973  & 0.968  & 0.964  & 0.971 & 0.965 & 0.985\\  

\bottomrule
\end{tabular}
\label{Table:CNNDetector_Similarity}
\end{table*}

\subsection{CNNDetector}
In \textbf{CNNDetector}, it is evaluated on a large number of 13 GAN-based image generation methods. In their original experiments, the objects in the images are very different, containing animals, human faces, road, \etc. For each GAN-based image generation method, the size of the testing dataset ranges from hundreds to thousands. 
They use detection accuracy and average precision (AP) as the metrics, on the same number of fake images and real images as the testing dataset.
We follow the same evaluation setting, except replacing fake images with reconstructed images by our methods. 

As we can see in Table \ref{Table:CNNDetector_Acc}, the two models used by \textbf{CNNDetector}: blur\_jpg\_prob0.1 (prob0.1) and blur\_jpg\_prob0.5 (prob0.5) achieve high accuracy. For convenience, we only introduce the data of using blur\_jpg\_prob0.1. In the first row, \textbf{Real \& Fake} means using real images and fake images as the testing dataset (\ie, the same testing dataset as used in \textbf{CNNDetector}). 

To show the detection accuracy of real images and fake images, we conduct extra experiments on real images and fake images respectively. As shown in the second and third row, the testing datasets are {Real} images only and {Fake} images only. In the second row, we can observe that the model performs well. However, in the third row, the performance of the model of \textbf{CNNDetector} is different in various GAN-based image generation methods. It has various performance drops on SAN and DeepFakes although it achieves high accuracy on GANs, \eg, ProGAN, StarGAN, CRN.  

For PCA reconstruction of our method, we produce a corresponding reconstructed image for each fake image in the testing dataset. For K-SVD reconstruction, it needs a lot of time to produce one K-SVD image. Thus for each category of each GAN-based image generation method, we choose 100 fake images and produce 100 corresponding K-SVD reconstructed images. No matter the detection accuracy of fake images, the reconstructed images generated by us can reduce its performance. 

As we can see in the fourth and fifth row, the testing datasets are PCA-reconstructed images and K-SVD-reconstructed images respectively. Compared to the detection accuracy of that on fake images, most of them are both decreased. Only the accuracy of SAN increases slightly. In the experiment of blur\_jpg\_prob0.5, all the detection accuracy decrease.
The similarity of PCA/K-SVD reconstructed images and fake images is shown in Table \ref{Table:CNNDetector_Similarity}. The images of CycleGAN, StyleGAN, StyleGAN2 and ProGAN have quite a few different categories. For these four multi-category GANs, they use different folders to store different categories of images. In the other nine GANs, some of them involve only one category (DeepFakes, IMLE, StarGAN, Whichfaceisreal, CRN). The others combine images of different categories into one folder. We call these GANs (BigGAN, GauGAN, SAN, SITD) uncertain-category and use `\textbf{-}' to represent the category.

\subsection{Comparison Between Partial and Full Reconstruction}
Sometimes, fake images are produced by only modifying parts of the real images. For example, DeepFake methods may change the hair color of a person or the color of a chair in the bedroom. For these situations, we propose partial reconstruction. 

In the GANs of \textbf{GANFingerprint}, \textbf{DCTA} and \textbf{CNNDetection}, only \textbf{StarGAN} have partially modified fake images. Therefore, we use StarGAN to show the comparison between partial reconstruction and full reconstruction. The numbers of real images and fake images of StarGAN are 1,999. We produce 1,999 reconstructed images for partial and full reconstruction of PCA. We also produce 100 reconstructed images for K-SVD. Since it needs a lot of time to produce K-SVD images, the number of K-SVD reconstructed images is not as large as that of PCA reconstructed images.

For each of \textbf{GANFingerprint} and \textbf{DCTA}, we train a binary classification model with real images from CelebA and partially modified fake images from StarGAN.

Table \ref{Table:Comparison_accuracy_partial_full} summarizes the detection accuracy of fully reconstructed images and partially reconstructed images on three different types of fake detectors. 
Compared with full reconstruction, the detection accuracy of all the three fake detectors decrease similarly in partial reconstruction for both PCA and K-SVD. Table \ref{Table:Comparison_similarity_partial_full}, shows the similarity metrics between reconstructed images and the original fake images. We can observe that the similarity of fake images and partial reconstruction is higher than that of fake images and full reconstruction.  

\begin{table}[tbp]
\footnotesize
\centering
\caption{Comparison of detection accuracy between partial reconstruction and full reconstruction}
\setlength{\tabcolsep}{0.7pt}
\begin{tabular}{c c c c c}
\toprule

{Accuracy(\%)} & GANFingerprint & DCTA & CNNDetection(0.1) &  CNNDetection(0.5)\\ 
\toprule
Fake                & 99.6  & 76.1 & 86.7 & 62.7\\ 
fully PCA   & 92.6  \textcolor{red}{(-7.0)}& 65.2 \textcolor{red}{(-10.9)}& 36.0 \textcolor{red}{(-50.7)}& 42.1 \textcolor{red}{(-20.6)}\\ 
partially PCA   & 92.5 \textcolor{red}{(-7.1)}&  64.7  \textcolor{red}{(-11.4)}& 26.0 \textcolor{red}{(-60.7)}& 36.6 \textcolor{red}{(-26.1)}\\ 
fully K-SVD  & 87.0 \textcolor{red}{(-12.6)}&  40.0 \textcolor{red}{(-36.1)}& 48.0 \textcolor{red}{(-38.7)}& 37.0  \textcolor{red}{(-25.7)}\\ 
partially K-SVD  & 87.0  \textcolor{red}{(-12.6)}& 38.9  \textcolor{red}{(-37.2)}& 52.0 \textcolor{red}{(-34.7)}& 35.0 \textcolor{red}{(-27.7)}\\ 

\bottomrule
\end{tabular}
\label{Table:Comparison_accuracy_partial_full}
\end{table}

\begin{table}[tbp]
\footnotesize
\centering
\caption{Similarity comparison between partial reconstruction and full reconstruction}
\setlength{\tabcolsep}{3.5pt}
\begin{tabular}{c c c c}
\toprule

{} &  COSS & PSNR & SSIM  \\ \toprule
fully PCA-reconstructed      &  0.99932 & 35.597 & 0.97072 \\
partially PCA-reconstructed  &  0.99936 & 35.221 & 0.97154 \\ 
fully K-SVD-reconstructed     &  0.99978 & 38.626 & 0.98713\\ 
partially K-SVD-reconstructed &  0.99978 & 38.623 & 0.98715\\

\bottomrule
\end{tabular}
\label{Table:Comparison_similarity_partial_full}
\end{table}

To sum up, compared with fully reconstructed images, the partially reconstructed images are more similar to their original fake counterparts.
Meanwhile, in terms of degrading the performance of fake detectors, their abilities are close, indicating the advantage of partial reconstruction for partially modified fake images.


\section{Conclusions}\label{sec:conc}
In the paper, we propose the \emph{FakePolisher}, a post-processing shallow reconstruction method based on dictionary learning without knowing any information of the GAN. 
The reconstructed images can easily fool the existing state-of-the-art detection methods. We also demonstrate that the existing detection methods are limited, which highly rely on the imperfection of upsampling methods. More powerful defense mechanisms for DeepFakes should be proposed. In future work, we plan to propose new methods that can remove the artifacts in fake images. Moreover, other shallow methods such as the ones based on advanced correlation filters \cite{icip16_paint,tip15_spartans,cvpr16_dikf,Zhou17_ICASSP,Zhang_neuro2019,ChenICME2018,DSARCF_TIP2019,Guo_TIP2020} are also potentially viable solutions to this problem, which we intend to explore further.



\balance

\bibliographystyle{ACM-Reference-Format}
\bibliography{ref}


\end{document}